\documentclass[11pt,a4paper]{article}
\usepackage[hyperref]{acl2017}
\usepackage{times}
\usepackage{latexsym}

\usepackage{url}

\usepackage{times}
\usepackage{latexsym}
\usepackage{graphicx}
\usepackage{amsmath}
\usepackage{multirow}
\graphicspath{ {Figure/} }
\usepackage{tabularx}
\usepackage{subcaption}
\usepackage{latexsym} 
\usepackage{xcolor}
\usepackage{pifont}

\aclfinalcopy 

\newcommand*{\affaddr}[1]{#1} 
\newcommand*{\affmark}[1][*]{\textsuperscript{#1}}
\newcommand*{\email}[1]{\texttt{#1}}

\title{Natural Language Generation for Spoken Dialogue System \\using RNN Encoder-Decoder Networks}

\author{%
Van-Khanh Tran\affmark[1,2] and Le-Minh Nguyen\affmark[1]\\
\affaddr{\affmark[1]Japan Advanced Institute of Science and Technology, JAIST\\
					1-1 Asahidai, Nomi, Ishikawa, 923-1292, Japan}\\
\email{\{tvkhanh, nguyenml\}@jaist.ac.jp}\\
\affaddr{\affmark[2]University of Information and Communication Technology, ICTU\\
	Thai Nguyen University, Vietnam}\\
\email{tvkhanh@ictu.edu.vn}
}

\begin{document}
\date{}
\maketitle

\begin{abstract}
Natural language generation (NLG) is a critical component in a spoken dialogue system. 
This paper presents a Recurrent Neural Network based Encoder-Decoder architecture, in which an LSTM-based decoder is introduced to select, aggregate semantic elements produced by an attention mechanism over the input elements, and to produce the required utterances.
The proposed generator can be jointly trained both sentence planning and surface realization to produce natural language sentences.
The proposed model was extensively evaluated on four different NLG datasets.
The experimental results showed that the proposed generators not only consistently outperform the previous methods across all the NLG domains but also show an ability to generalize from a new, unseen domain and learn from multi-domain datasets. 
\end{abstract}

\section{Introduction}\label{sec:introduction}
Natural Language Generation (NLG) plays a critical role in Spoken Dialogue Systems (SDS) with task is to convert a meaning representation produced by the Dialogue Manager into natural language utterances. Conventional approaches still rely on comprehensive hand-tuning templates and rules requiring expert knowledge of linguistic representation, including rule-based \cite{mirkovic2011dialogue}, corpus-based n-gram models \cite{oh2000stochastic}, and a trainable generator \cite{stent2004trainable}.

Recently, Recurrent Neural Networks (RNNs) based approaches have shown promising performance in tackling the NLG problems. The RNN-based models have been applied for NLG as a joint training model \cite{thwsjy15,wensclstm15} and an end-to-end training model \cite{wen2016network}. A recurring problem in such systems is requiring annotated datasets for particular dialogue acts\footnote{A combination of an action type and a set of slot-value pairs. e.g. \textit{inform(name='Bar crudo'; food='raw food')}} (DAs). To ensure that the generated utterance representing the intended meaning of the given DA, the previous RNN-based models were further conditioned on a 1-hot vector representation of the DA. \citet{thwsjy15} introduced a heuristic gate to ensure that all the slot-value pair was accurately captured during generation. \citet{wensclstm15} subsequently proposed a Semantically Conditioned Long Short-term Memory generator (SC-LSTM) which jointly learned the DA gating signal and language model.

More recently, Encoder-Decoder networks \cite{vinyals2015neural,li2015diversity}, especially the attentional based models \cite{wentoward,mei2015talk} have been explored to solve the NLG tasks. The Attentional RNN Encoder-Decoder \cite{bahdanau2014neural} (ARED) based approaches have also shown improved performance on a variety of tasks, e.g., image captioning \cite{xu2015show,yang2016review}, text summarization \cite{rush2015neural,nallapati2016abstractive}.


While the RNN-based generators with DA gating-vector can prevent the undesirable semantic repetitions, the ARED-based generators show signs of better adapting to a new domain. However, none of the models show significant advantage from out-of-domain data. To better analyze model generalization to an unseen, new domain as well as model leveraging the out-of-domain sources, we propose a new architecture which is an extension of the ARED model. 
In order to better select, aggregate and control the semantic information, a Refinement Adjustment LSTM-based component (\textit{RALSTM}) is introduced to the decoder side. The proposed model can learn from unaligned data by jointly training the sentence planning and surface realization to produce natural language sentences. We conducted experiments on four different NLG domains and found that the proposed methods significantly outperformed the state-of-the-art methods regarding BLEU \cite{papineni2002bleu} and slot error rate ERR scores \cite{wensclstm15}. The results also showed that our generators could scale to new domains by leveraging the out-of-domain data. To sum up, we make three key contributions in this paper:
\begin{itemize}
\item We present an LSTM-based component called \textit{RALSTM} cell applied on the decoder side of an ARED model, resulting in an end-to-end generator that empirically shows significant improved performances in comparison with the previous approaches.
\item We extensively conduct the experiments to evaluate the models training from scratch on each in-domain dataset.
\item We empirically assess the models' ability to: learn from multi-domain datasets by pooling all available training datasets; and adapt to a new, unseen domain by limited feeding amount of in-domain data.
\end{itemize}
We review related works in Section \ref{sec:relatedwork}. Following a detail of proposed model in Section \ref{sec:method}, Section \ref{sec:experiments} describes datasets, experimental setups, and evaluation metrics. The resulting analysis is presented in Section \ref{sec:resultsandanalysis}. We conclude with a brief summary and future work in Section \ref{sec:conclusion}.
\section{Related Work}\label{sec:relatedwork}
Recently, RNNs-based models have shown promising performance in tackling the NLG problems. \citet{zhang2014chinese} proposed a generator using RNNs to create Chinese poetry. \citet{xu2015show,karpathy2015deep,vinyals2015show} also used RNNs in a multi-modal setting to solve image captioning tasks. The RNN-based Sequence to Sequence models have applied to solve variety of tasks: conversational modeling \cite{vinyals2015neural,li2015diversity,li2016persona}, machine translation \cite{luong2015multi,li2016mutual}

For task-oriented dialogue systems, \citet{thwsjy15} combined a forward RNN generator, a CNN reranker, and a backward RNN reranker to generate utterances. \citet{wensclstm15} proposed SC-LSTM generator which introduced a control sigmoid gate to the LSTM cell to jointly learn the gating mechanism and language model. A recurring problem in such systems is the lack of sufficient domain-specific annotated data. \citet{wen2016multi} proposed an out-of-domain model which was trained on counterfeited data by using semantically similar slots from the target domain instead of the slots belonging to the out-of-domain dataset. The results showed that the model can achieve a satisfactory performance with a small amount of in-domain data by fine tuning the target domain on the out-of-domain trained model. 

More recently, RNN encoder-decoder based models with attention mechanism \cite{bahdanau2014neural} have shown improved performances in various tasks. \citet{yang2016review} proposed a review network to the image captioning, which reviews all the information encoded by the encoder and produces a compact thought vector. \citet{mei2015talk} proposed RNN encoder-decoder-based model by using two attention layers to jointly train content selection and surface realization. More close to our work, \citet{wentoward} proposed an attentive encoder-decoder based generator which computed the attention mechanism over the slot-value pairs. The model showed a domain scalability when a very limited amount of data is available.

Moving from a limited domain dialogue system to an open domain dialogue system raises some issues. Therefore, it is important to build an open domain dialogue system that can make as much use of existing abilities of functioning from other domains. There have been several works to tackle this problem, such as \cite{mrkvsic2015multi} using RNN-based networks for multi-domain dialogue state tracking, \cite{wen2016multi} using a procedure to train multi-domain via multiple adaptation steps, or \cite{gavsic2015distributed,williams2013multi} adapting of SDS components to new domains. 
\section{Recurrent Neural Language Generator}\label{sec:method}
\begin{figure}[!ht]
	\centering
    \includegraphics[width=0.43\textwidth, height=.45\textwidth]{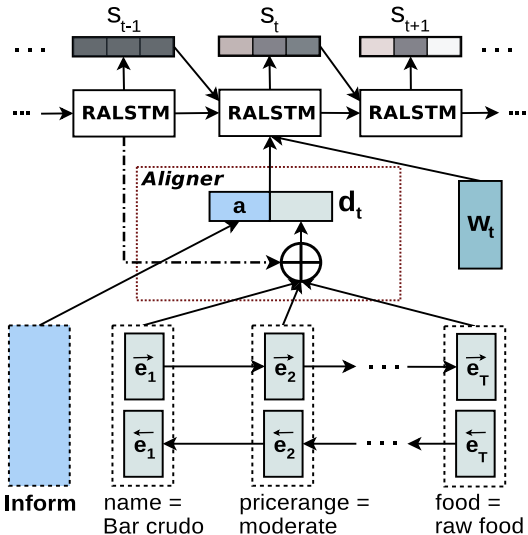}
    \caption{Unrolled presentation of the RNNs-based neural language generator. The Encoder part is a BiLSTM, the Aligner is an attention mechanism over the encoded inputs, and the Decoder is the proposed RALSTM model conditioned on a 1-hot representation vector $\textbf{s}$. The fading color of the vector $\textbf{s}$ indicates retaining information for future computational time steps.}
    \label{fig:nlg-model}
\end{figure}
The recurrent language generator proposed in this paper is based on a neural language generator \cite{wentoward}, which consists of three main components: (i) an Encoder that incorporates the target meaning representation (MR) as the model inputs, (ii) an Aligner that aligns and controls the semantic elements, and (iii) an RNN Decoder that generates output sentences. The generator architecture is shown in Figure \ref{fig:nlg-model}. The Encoder first encodes the MR into input semantic elements which are then aggregated and selected by utilizing an attention-based mechanism by the Aligner. The input to the RNN Decoder at each time step is a 1-hot encoding of a token\footnote{Input texts are delexicalized where slot values are replaced by its corresponding slot tokens.} $\textbf{w}_{t}$ and an attentive DA representation $\textbf{d}_{t}$. At each time step $t$, RNN Decoder also computes how much the feature value vector $\textbf{s}_{t-1}$ retained for the next computational steps, and adds this information to the RNN output which represents the probability distribution of the next token $\textbf{w}_{t+1}$. At generation time, we can sample from this conditional distribution to obtain the next token in a generated sentence, and feed it as the next input to the RNN Decoder. This process finishes when an end sign is generated \cite{karpathy2015deep}, or some constraints are reached \cite{zhang2014chinese}. The model can produce a sequence of tokens which can finally be lexicalized\footnote{The process in which slot token is replaced by its value.} to form the required utterance. 

\begin{figure}[!ht]
	\centering
    \includegraphics[width=0.45\textwidth, height=.45\textwidth]{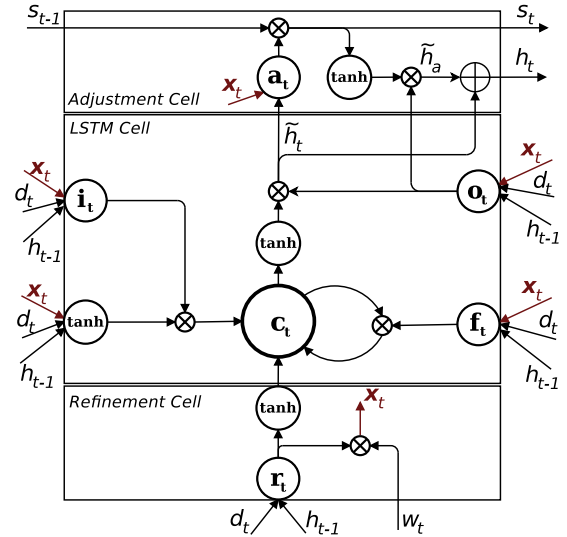}
    \caption{The RALSTM cell proposed in this paper, which consists of three components: an Refinement Cell, a traditional LSTM Cell, and an Adjustment Cell. At time step $t$, while the Refinement cell computes new input tokens $\textbf{x}_{t}$ based on the original input tokens and the attentional DA representation $\textbf{d}_{t}$, the Adjustment Cell calculates how much information of the slot-value pairs can be generated by the LSTM Cell. }
    \label{fig:RALSTM-model}
\end{figure}

\subsection{Encoder}
The slots and values are separated parameters used in the encoder side. This embeds the source information into a vector representation $\textbf{z}_{i}$ which is a concatenation of embedding vector representation of each slot-value pair, and is computed by:
\begin{equation}\label{eq:z-i-1}
\textbf{z}_{i} = \textbf{u}_{i} \oplus \textbf{v}_{i}
\end{equation}
where $\textbf{u}_{i}$, $\textbf{v}_{i}$ are the $i$-th slot and value embedding vectors, respectively, and $\oplus$ is vector concatenation. The \textit{i} index runs over the $L$ given slot-value pairs.
In this work, we use a 1-layer, Bidirectional LSTM (Bi-LSTM) to encode the sequence of slot-value pairs\footnote{We treated the set of slot-value pairs as a sequence and use the order specified by slot's name (e.g., slot \textit{address} comes first, \textit{food} follows \textit{address}). We have tried treating slot-value pairs as a set with natural order as in the given DAs. However, this yielded even worse results.} embedding. The Bi-LSTM consists of forward and backward LSTMs which read the sequence of slot-value pairs from left-to-right and right-to-left to produce forward and backward sequence of hidden states ($\overrightarrow{\textbf{e}_{1}}, .., \overrightarrow{\textbf{e}_{L}}$), and  ($\overleftarrow{\textbf{e}_{1}}, .., \overleftarrow{\textbf{e}_{L}}$), respectively. We then obtain the sequence of encoded hidden states $\textbf{E}=(\textbf{e}_{1}, \textbf{e}_{2}, .., \textbf{e}_{L})$ where $\textbf{\textbf{e}}_{i}$ is a sum of the forward hidden state $\overrightarrow{\textbf{e}_{i}}$ and the backward one $\overleftarrow{\textbf{e}_{i}}$ as follows:
\begin{equation}\label{eq:b-i}
\textbf{e}_{i}=\overrightarrow{\textbf{e}_{i}} + \overleftarrow{\textbf{e}_{i}}
\end{equation}
\subsection{Aligner}\label{subsec:aligner}
The Aligner utilizes attention mechanism to calculate the DA representation as follows:
\begin{equation}
\beta_{t,i} = \frac{\exp e_{t,i} }{\sum\nolimits_{j}\exp e_{t,j}}
\end{equation}
where
\begin{equation}
e_{t,i}=a(\textbf{e}_{i}, \textbf{h}_{t-1})
\end{equation}
and  $\beta_{t,i}$ is the weight of \textit{i}-th slot-value pair calculated by the attention mechanism. The alignment model $a$ is computed by:
\begin{equation}\label{eq:alignment-a}
a(\textbf{e}_{i}, \textbf{h}_{t-1}) = \textbf{v}_{a}^{\top}\tanh(\textbf{W}_{a}\textbf{e}_{i} + \textbf{U}_{a}\textbf{h}_{t-1})
\end{equation}
where $\textbf{v}_{a}, \textbf{W}_{a}, \textbf{U}_{a}$ are the weight matrices to learn.
Finally, the Aligner calculates dialogue act embedding $\textbf{d}_{t}$ as follows:
\begin{equation}\label{eq:d-t}
\textbf{d}_{t} = \textbf{a} \oplus \sum\nolimits_{i}\beta_{t,i} \textbf{e}_{i}
\end{equation} 
where \textbf{a} is vector embedding of the action type.

\subsection{RALSTM Decoder}\label{subsec:ralstm}
The proposed semantic RALSTM cell applied for Decoder side consists of three components: a Refinement cell, a traditional LSTM cell, and an Adjustment cell:

Firstly, instead of feeding the original input token $\textbf{w}_{t}$ into the RNN cell, the input is recomputed by using a semantic gate as follows:
\begin{equation}\label{eq:d-t-1}
\begin{aligned}
	\textbf{r}_{t}&=\sigma(\textbf{W}_{rd}\textbf{d}_{t} + \textbf{W}_{rh}\textbf{h}_{t-1})\\
	\textbf{x}_{t}&=\textbf{r}_{t} \odot \textbf{w}_{t} 
\end{aligned}
\end{equation}
where $\textbf{W}_{rd}$ and $\textbf{W}_{rh}$ are weight matrices. Element-wise multiplication $\odot$ plays a part in word-level matching which not only learns the vector similarity, but also preserves information about the two vectors. $\textbf{W}_{rh}$ acts like a key phrase detector that learns to capture the pattern of generation tokens or the relationship between multiple tokens. In other words, the new input $\textbf{x}_{t}$ consists of information of the original input token $\textbf{w}_{t}$, the DA representation $\textbf{d}_{t}$, and the hidden context $\textbf{h}_{t-1}$. $\textbf{r}_{t}$ is called a \textit{Refinement} gate because the input tokens are refined by a combination gating information of the attentive DA representation $\textbf{d}_{t}$ and the previous hidden state $\textbf{h}_{t-1}$. By this way, we can represent the whole sentence based on the refined inputs.

Secondly, the traditional LSTM network proposed by \citeauthor{hochreiter1997long}~\shortcite{bahdanau2014neural} in which the input gate $\textbf{i}_{i}$, forget gate $\textbf{f}_{t}$ and output gates $\textbf{o}_{t}$ are introduced to control information flow and computed as follows:
\begin{equation}\label{eq:lstm}
\begin{aligned}
\begin{pmatrix}
\textbf{i}_{t}
\\ \textbf{f}_{t}
\\ \textbf{o}_{t}
\\ \hat{\textbf{c}}_{t}
\end{pmatrix}
&=
\begin{pmatrix}\sigma
\\ \sigma
\\ \sigma
\\ \tanh
\end{pmatrix}\textbf{W}_{4n,4n}
\begin{pmatrix}
\textbf{x}_{t}
\\ \textbf{d}_{t}
\\ \textbf{h}_{t-1}
\end{pmatrix}\\
\end{aligned}
\end{equation}
where $n$ is hidden layer size, $\textbf{W}_{4n,4n}$ is model parameters. The cell memory value $\textbf{c}_{t}$ is modified to depend on the DA representation as:
\begin{equation}\label{eq:lstm-c-o-t}
\begin{aligned}
\textbf{c}_{t}&=\textbf{f}_{t}\odot\textbf{c}_{t-1} +\textbf{i}_{t}\odot \hat{\textbf{c}}_{t} + \tanh(\textbf{W}_{cr}\textbf{r}_{t})
\\ \tilde{\textbf{h}}_{t}&= \textbf{o}_{t} \odot \tanh(\textbf{c}_{t})
\end{aligned}
\end{equation}
where $\tilde{\textbf{h}}_{t}$ is the output.

Thirdly, inspired by work of \citet{wensclstm15} in which the generator was further conditioned on a 1-hot representation vector $\textbf{s}$ of given dialogue act, and work of \citet{lu2016knowing} that proposed a visual sentinel gate to make a decision on whether the model should attend to the image or to the sentinel gate, an additional gating cell is introduced on top of the traditional LSTM to gate another controlling vector $\textbf{s}$. Figure \ref{fig:DAVectoControlling} shows how RALSTM controls the DA vector $\textbf{s}$. First, starting from the 1-hot vector of the DA $\textbf{s}_{0}$, at each time step $t$ the proposed cell computes how much the LSTM output $\tilde{\textbf{h}}_{t}$ affects the DA vector, which is computed as follows:
\begin{equation}\label{eq:a-t-1}
\begin{aligned}
\textbf{a}_{t}&=\sigma(\textbf{W}_{ax}\textbf{x}_{t} +\textbf{W}_{ah}\tilde{\textbf{h}}_{t})\\
\textbf{s}_{t}&=\textbf{s}_{t-1} \odot \textbf{a}_{t}
\end{aligned}
\end{equation}
where $\textbf{W}_{ax}$, $\textbf{W}_{ah}$ are weight matrices to be learned. $\textbf{a}_{t}$ is called an $Adjustment$ gate since its task is to control what information of the given DA have been generated and what information should be retained for future time steps. Second, we consider how much the information preserved in the DA $\textbf{s}_{t}$ can be contributed to the output, in which an additional output is computed by applying the output gate $\textbf{o}_{t}$ on the remaining information in $\textbf{s}_{t}$ as follows:
\begin{equation}\label{eq:h-a-1}
\begin{aligned}
\textbf{c}_{a}&=\sigma(\textbf{W}_{os}\textbf{s}_{t})\\
\tilde{\textbf{h}}_{a}&= \textbf{o}_{t} \odot \tanh(\textbf{c}_{a})
\end{aligned}
\end{equation}
where $\textbf{W}_{os}$ is a weight matrix to project the DA presentation into the output space, $\tilde{\textbf{h}}_{a}$ is the Adjustment cell output. Final RALSTM output is a combination of both outputs of the traditional LSTM cell and the Adjustment cell, and computed as follows:
\begin{equation}\label{eq:h-t-1}
\textbf{h}_{t}=\tilde{\textbf{h}}_{t} + \tilde{\textbf{h}}_{a}
\end{equation}

Finally, the output distribution is computed by applying a softmax function $g$, and the distribution can be sampled to obtain the next token,
\begin{equation}\label{eq:p-t-1}
\begin{aligned}
& P(w_{t+1}\mid w_{t},...w_{0},\textbf{DA})=g(\textbf{W}_{ho}\textbf{h}_{t}) \\
& w_{t+1} \sim P(w_{t+1}\mid w_{t}, w_{t-1},...w_{0},\textbf{DA})
\end{aligned}
\end{equation}
where $\textbf{DA}=(\textbf{s}, \textbf{z})$.
\subsection{Training}\label{subsec:training}
The objective function was the negative log-likelihood and computed by:
\begin{equation}\label{eq:c-f-1}
\textbf{F}(\theta) = -\sum_{t=1}^{T}\textbf{y}_{t}^{\top}\log{\textbf{p}_{t}}
\end{equation}
where: $\textbf{y}_{t}$ is the ground truth token distribution, $\textbf{p}_{t}$ is the predicted token distribution, $T$ is length of the input sentence. The proposed generators were trained by treating each sentence as a mini-batch with \textit{$l_{2}$} regularization added to the objective function for every 5 training examples. The models were initialized with a pretrained Glove word embedding vectors \cite{pennington2014glove} and optimized by using stochastic gradient descent and back propagation through time \cite{werbos1990backpropagation}. Early stopping mechanism was implemented to prevent over-fitting by using a validation set as suggested in \cite{mikolov2010recurrent}.

\subsection{Decoding}\label{subsec:decoding}
The decoding consists of two phases: (i) over-generation, and (ii) reranking. In the over-generation, the generator conditioned on both representations of the given DA use a beam search to generate a set of candidate responses. In the reranking phase, cost of the generator is computed to form the reranking score $\textbf{R}$ as follows:
\begin{equation}\label{eq:r-score-1}
\textbf{R} = \textbf{F}(\theta) + \lambda \textbf{ERR}
\end{equation}
where $\lambda$ is a trade off constant and is set to a large value in order to severely penalize nonsensical outputs. The slot error rate $\textbf{ERR}$, which is the number of slots generated that is either missing or redundant, and is computed by:
\begin{equation}
\textbf{ERR} = \frac{\textbf{p} + \textbf{q}}{\textbf{N}}
\end{equation}
where $\textbf{N}$ is the total number of slots in DA, and $\textbf{p}$, $\textbf{q}$ is the number of missing and redundant slots, respectively. 
\section{Experiments}\label{sec:experiments}
We extensively conducted a set of experiments to assess the effectiveness of the proposed models by using several metrics, datasets, and model architectures, in order to compare to prior methods.
\subsection{Datasets}\label{subsec:datasets}
We assessed the proposed models on four different NLG domains: finding a restaurant, finding a hotel, buying a laptop, and buying a television. The Restaurant and Hotel were collected in \cite{wensclstm15}, while the Laptop and TV datasets have been released by \cite{wen2016multi} with a much larger input space but only one training example for each DA so that the system must learn partial realization of concepts and be able to recombine and apply them to unseen DAs. This makes the NLG tasks for the Laptop and TV domains become much harder. The dataset statistics are shown in Table \ref{tab:tab-dataset-stat}.
\begin{table}[!ht]
\centering
\caption{Dataset statistics.}
\label{tab:tab-dataset-stat}
\scalebox{0.8}{
\begin{tabular}{ccccc}
\hline
              & \textbf{Restaurant} & \textbf{Hotel} & \textbf{Laptop} & \textbf{TV} \\ \hline
\# train      & 3,114                & 3,223 			 & 7,944 		   & 4,221 \\ 
\# validation & 1,039                & 1,075 			 & 2,649 		   & 1,407 \\
\# test       & 1,039 				& 1,075 			 & 2,649 		   & 1,407 \\
\# distinct DAs   & 248 			& 164 			 & 13,242		   & 7,035 \\
\# DA types   & 8 				    & 8 			 & 14		       & 14 \\
\# slots 	  & 12 				    & 12 		 	 & 19 		   & 15 \\
\hline
\end{tabular}
}
\end{table}
\begin{table*}[!ht]
\centering
\caption{Performance comparison on four datasets in terms of the BLEU and the error rate ERR(\%) scores. The results were produced by training each network on 5 random initialization and selected model with the highest validation BLEU score. $^{\sharp}$ denotes the Attention-based Encoder-Decoder model. The best and second best models highlighted in \textbf{bold} and \textbf{\textit{italic}} face, respectively.}
\label{tab:tab-performance}
\scalebox{0.9}{
\begin{tabular}{ccccccccc}
\hline 
\multirow{2}{*}{Model} & \multicolumn{2}{c}{\textbf{Restaurant}} & \multicolumn{2}{c}{\textbf{Hotel}} & \multicolumn{2}{c}{\textbf{Laptop}} & \multicolumn{2}{c}{\textbf{TV}} \\ \cline{2-9} 
 & BLEU & ERR & BLEU & ERR & BLEU & ERR & BLEU & ERR \\ \hline
HLSTM & 0.7466 & 0.74\% & 0.8504 & 2.67\% & 0.5134 & 1.10\% & 0.5250 & 2.50\% \\
SCLSTM & 0.7525 & 0.38\% & 0.8482 & 3.07\% & 0.5116 & 0.79\% & 0.5265 & 2.31\% \\
Enc-Dec$^{\sharp}$ & 0.7398 & 2.78\% & 0.8549 & 4.69\% & 0.5108 & 4.04\% & 0.5182 & 3.18\% \\ \hline \hline
w/o A$^{\sharp}$  & 0.7651 & 0.99\% & 0.8940 & 1.82\% & 0.5219& 1.64\% & 0.5296 & 2.40\% \\
w/o R$^{\sharp}$	& \textbf{\textit{0.7748}} & \textbf{\textit{0.22}}\% & \textit{\textbf{0.8944}} & \textbf{\textit{0.48}}\% & \textbf{\textit{0.5235}}& \textbf{\textit{0.57}}\% & \textbf{\textit{0.5350}} & \textbf{\textit{0.72}}\% \\ 
RALSTM$^{\sharp}$	& \textbf{0.7789} & \textbf{0.16}\% & \textbf{0.8981}
 & \textbf{0.43}\% & \textbf{0.5252} & \textbf{0.42}\% & \textbf{0.5406} & \textbf{0.63}\%
\end{tabular}
}
\end{table*}

\begin{table*}[!ht]
\centering
\caption{Performance comparison of the proposed models on four datasets in terms of the BLEU and the error rate ERR(\%) scores. The results were averaged over 5 randomly initialized networks. \textbf{bold} denotes the best model.}
\label{tab:tab-average-performance}
\scalebox{0.9}{
\begin{tabular}{ccccccccc}
\hline
\multirow{2}{*}{Model} & \multicolumn{2}{c}{\textbf{Restaurant}} & \multicolumn{2}{c}{\textbf{Hotel}} & \multicolumn{2}{c}{\textbf{Laptop}} & \multicolumn{2}{c}{\textbf{TV}} \\ \cline{2-9} 
 & BLEU & ERR & BLEU & ERR & BLEU & ERR & BLEU & ERR \\ \hline
w/o A & 0.7619 & 2.26\% & 0.8913 & 1.85\% & 0.5180& 1.81\% & 0.5270 & 2.10\% \\ 
w/o R & 0.7733 & 0.23\% & 0.8901 & 0.59\% & 0.5208 & 0.60\% & 0.5321 & 0.50\% \\
RALSTM & \textbf{0.7779} & \textbf{0.20}\% & \textbf{0.8965} & \textbf{0.58}\% & \textbf{0.5231}& \textbf{0.50}\% & \textbf{0.5373} & \textbf{0.49}\% \\
\end{tabular}
}
\end{table*}
\subsection{Experimental Setups}\label{subsec:experimental-setups}
\begin{figure*}[!ht]
	\centering 
    \includegraphics[width=0.9\textwidth, height=3.5cm]{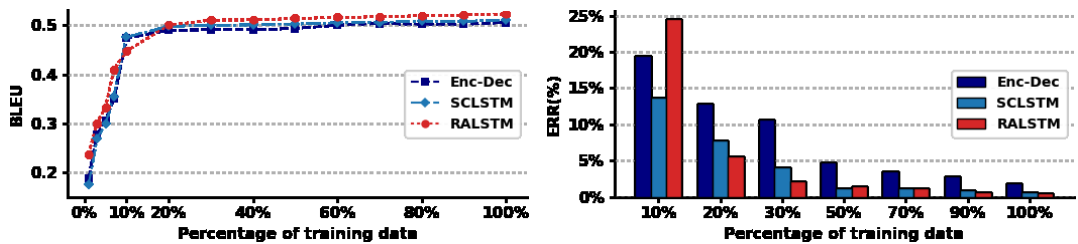}
    \caption{Performance comparison of the models trained on Laptop domain.}
    \label{fig:laptop-performances}
\end{figure*}

\begin{figure*}[!ht]
	\centering 
    \includegraphics[width=0.9\textwidth, height=3.5cm]{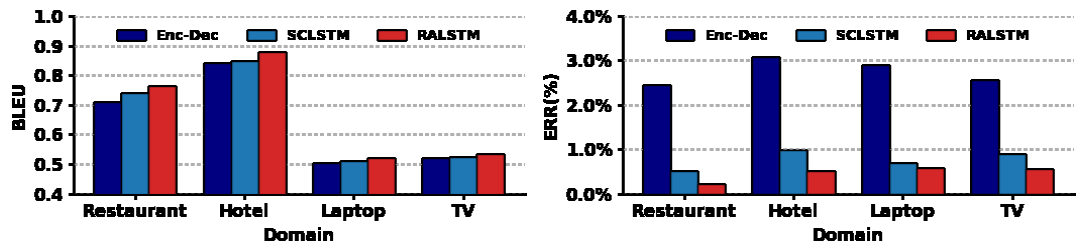}
    \caption{Performance comparison of the general models on four different domains.}
    \label{fig:general-models}
\end{figure*}

\begin{figure*}[!ht]
	\centering 
    \includegraphics[width=0.9\textwidth, height=3.5cm]{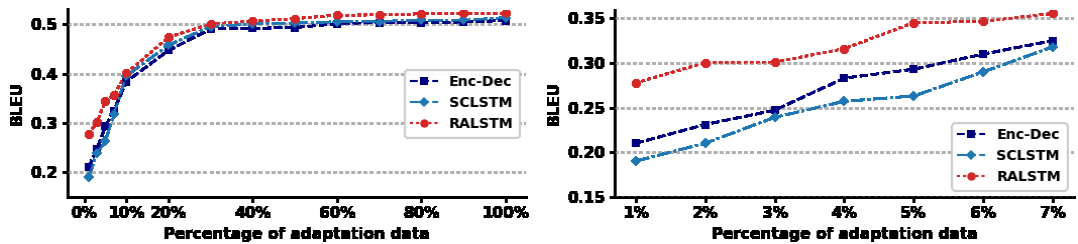}
    \caption{Performance on Laptop domain with varied amount of the adaptation training data when adapting models trained on Restaurant+Hotel dataset.}
    \label{fig:adaptation-models}
\end{figure*}

The generators were implemented using the TensorFlow library \cite{abadi2016tensorflow} and trained with training, validation and testing ratio as 3:1:1. The hidden layer size, beam size were set to be 80 and 10, respectively, and the generators were trained with a $70\%$ of dropout rate. We performed 5 runs with different random initialization of the network and the training is terminated by using early stopping. We then chose a model that yields the highest BLEU score on the validation set as shown in Table \ref{tab:tab-performance}. Since the trained models can differ depending on the initialization, we also report the results which were averaged over 5 randomly initialized networks. Note that, except the results reported in Table \ref{tab:tab-performance}, all the results shown were averaged over 5 randomly initialized networks. We set $\lambda$ to 1000 to severely discourage the reranker from selecting utterances which contain either redundant or missing slots. For each DA, we over-generated 20 candidate sentences and selected the top 5 realizations after reranking. Moreover, in order to better understand the effectiveness of our proposed methods, we: (i) performed an ablation experiments to demonstrate the contribution of each proposed cells (Tables \ref{tab:tab-performance}, \ref{tab:tab-average-performance}), (ii) trained the models on the Laptop domain with varied proportion of training data, starting from $10\%$ to $100\%$ (Figure \ref{fig:laptop-performances}), (iii) trained general models by merging all the data from four domains together and tested them in each individual domain (Figure \ref{fig:general-models}), and (iv) trained adaptation models on merging data from restaurant and hotel domains, then fine tuned the model on laptop domain with varied amount of adaptation data (Figure \ref{fig:adaptation-models}).

\subsection{Evaluation Metrics and Baselines}\label{subsec:evaluation-metrics}
The generator performance was assessed on the two evaluation metrics: the BLEU and the slot error rate ERR by adopting code from an open source benchmark toolkit for Natural Language Generation\footnotemark. We compared the proposed models against three strong baselines which have been recently published as state-of-the-art NLG benchmarks\footnotemark[\value{footnote}].
\footnotetext{https://github.com/shawnwun/RNNLG}
\begin{itemize}
    \item HLSTM proposed by \citeauthor{thwsjy15}~\shortcite{thwsjy15} which used a heuristic gate to ensure that all of the slot-value information was accurately captured when generating.
	\item SCLSTM proposed by \citeauthor{wensclstm15}~\shortcite{wensclstm15} which can jointly learn the gating signal and language model.
    \item Enc-Dec proposed by \citeauthor{wentoward}~\shortcite{wentoward} which applied the attention-based encoder-decoder architecture.
\end{itemize}
\section{Results and Analysis}\label{sec:resultsandanalysis}
\subsection{Results}\label{subsec:results}
We conducted extensive experiments on our models and compared against the previous methods. Overall, the proposed models consistently achieve the better performance regarding both evaluation metrics across all domains in all test cases. 
\begin{table*}[ht]
\centering
\caption{Comparison of top responses generated for some input dialogue acts between different models. Errors are marked in color (\textcolor{red}{missing}, \textcolor{blue}{misplaced} information). \textit{All2*} are general models.}
\label{tab:comparison}
\resizebox{\textwidth}{!}{%
\begin{tabularx}{1.3\textwidth}{llX}
\textbf{\#} & \textbf{Model} & \textbf{Generated Responses from Laptop Domain} \\ \hline 
\textit{1} & \textbf{\textit{Input DA}} & ?compare(name="satellite notus 19", pricerange="budget", drive="500 gb", name="portege thanatos 98", pricerange="expensive", drive="128 gb") \\
& \textbf{\textit{Reference}} & the satellite notus 19 has a 500 gb drive and is in the budget price range . on the other hand the portege thanatos 98 has a 128 gb drive and is in the expensive price range . which would you prefer
\\
& HLSTM & the satellite notus 19 is a budget priced laptop with a 500 gb drive and a \textcolor{blue}{128 gb drive} . which one do you prefer [\textcolor{red}{portege thanatos 98}, \textcolor{red}{expensive}] \\
& Enc-Dec & the satellite notus 19 is in the budget price range . the portege thanatos 98 has a \textcolor{blue}{500 gb drive} and is in the expensive price range . which one do you prefer [\textcolor{red}{128 gb}]\\
& SCLSTM & the satellite notus 19 is in the budget price range and has a 500 gb drive and the portege thanatos 98 has a 128 gb drive . which one do you prefer [\textcolor{red}{expensive}] \\
& RALSTM & the satellite notus 19 is in the budget price range with a 500 gb drive , while the portege thanatos 98 is in the expensive price range and has a 128 gb drive . which one do you prefer \\ 
& All2Laptop& the satellite notus 19 is a budget price range with a 500 gb drive and the portege thanatos 98 has a 128 gb drive and is in the expensive price range . which one do you prefer \\
\hline
\textbf{\#} & \textbf{Model} & \textbf{Generated Responses from TV Domain} \\ \hline 
\textit{2} & \textbf{\textit{Input DA}} & inform\_count(count="73", type="television", hasusbport="dontcare", hdmiport="2", screensizerange="dontcare") \\
& \textbf{\textit{Reference}} & there are 73 televisions if you do not care about whether it has usb port or not, with 2 hdmi ports and if you do not care about the screen size    \\
& HLSTM & there are 73 televisions with 2 hdmi ports and if you do not care about the number of usb ports [\textcolor{red}{screensizerange}] \\
& Enc-Dec & there are 73 televisions with 2 hdmi ports if you do not care about usb ports [\textcolor{red}{screensizerange}] \\
& SCLSTM & there are 73 televisions with 2 hdmi ports if you do not care about the screen size and if you do not care about the number of usb ports \\
& RALSTM & there are 73 televisions with 2 hdmi ports if you do not care about screen size or usb ports \\ 
& All2Tv & there are 73 televisions with 2 hdmi ports if you do not care about screen size or usb ports \\
\end{tabularx}%
}
\end{table*}
\subsubsection*{Model Comparison in an Unseen Domain}

The ablation studies (Tables \ref{tab:tab-performance}, \ref{tab:tab-average-performance}) demonstrate the contribution of different model components in which the models were assessed without Adjustment cell (\textit{w/o A}), or without Refinement cell (\textit{w/o R}). It clearly sees that the Adjustment cell contributes to reducing the slot error rate ERR score since it can effectively prevent the undesirable slot-value pair repetitions by gating the DA vector $\textbf{s}$. A comparison between the ARED-based models (denoted by $^{\sharp}$ in Table \ref{tab:tab-performance}) shows that the proposed models not only have better performance with higher the BLEU score but also significantly reduce the slot error rate ERR score by a large margin about $2\%$ to $4\%$ in every datasets. Moreover, a comparison between the models with gating the DA vector also indicates that the proposed models (\textit{w/o R}, \textit{RALSTM}) have significant improved performance on both the evaluation metrics across the four domains compared to the \textit{SCLSTM} model. The \textit{RALSTM} cell without the Refinement cell is similar as the \textit{SCLSTM} cell. However, it obtained the results much better than the SCLSTM baselines. This stipulates the necessary of the LSTM encoder and the Aligner in effectively partial learning the correlated order between slot-value representation in the DAs, especially for the unseen domain where there is only one training example for each DA.
Table \ref{tab:tab-average-performance} further demonstrates the stable strength of our models since the results' pattern stays unchanged compared to those in Table \ref{tab:tab-performance}. 

Figure \ref{fig:laptop-performances} shows a comparison of three models (\textit{Enc-Dec}, \textit{SCLSTM}, and \textit{RALSTM}) which were trained from scratch on the unseen laptop domain in varied proportion of training data, from $1\%$ to $100\%$. It clearly shows that the \textit{RALSTM} outperforms the previous models in all cases, while the \textit{Enc-Dec} has a much greater ERR score comparing to the two models. 

A comparison of top responses generated for some input DAs between different models are shown in Table \ref{tab:comparison}. While the previous models still produce some errors (missing and misplaced information), the proposed models (RALSTM and the models \textit{All2*} trained by pooling all datasets together) can generate appropriate sentences. We also found that the proposed models tend to generate more complete and concise sentences than the other models. 

\begin{figure}[!ht]
	\centering
    \begin{subfigure}[b]{0.48\textwidth}
      \centering
      \includegraphics[width=\textwidth, height=3.3cm]{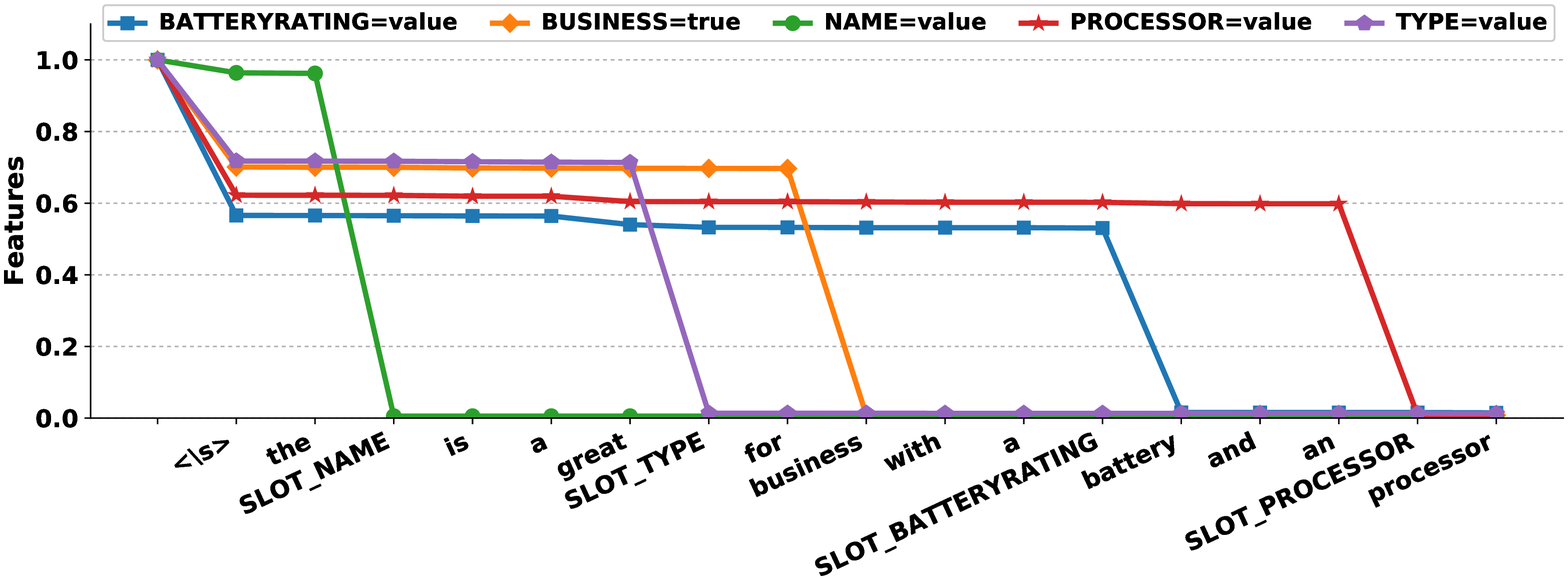}
      \subcaption{An example from the Laptop domain.}
      \label{fig:LaptopSample}
	\end{subfigure}
    \begin{subfigure}[b]{0.48\textwidth}
      \centering
      \includegraphics[width=\textwidth, height=3.3cm]{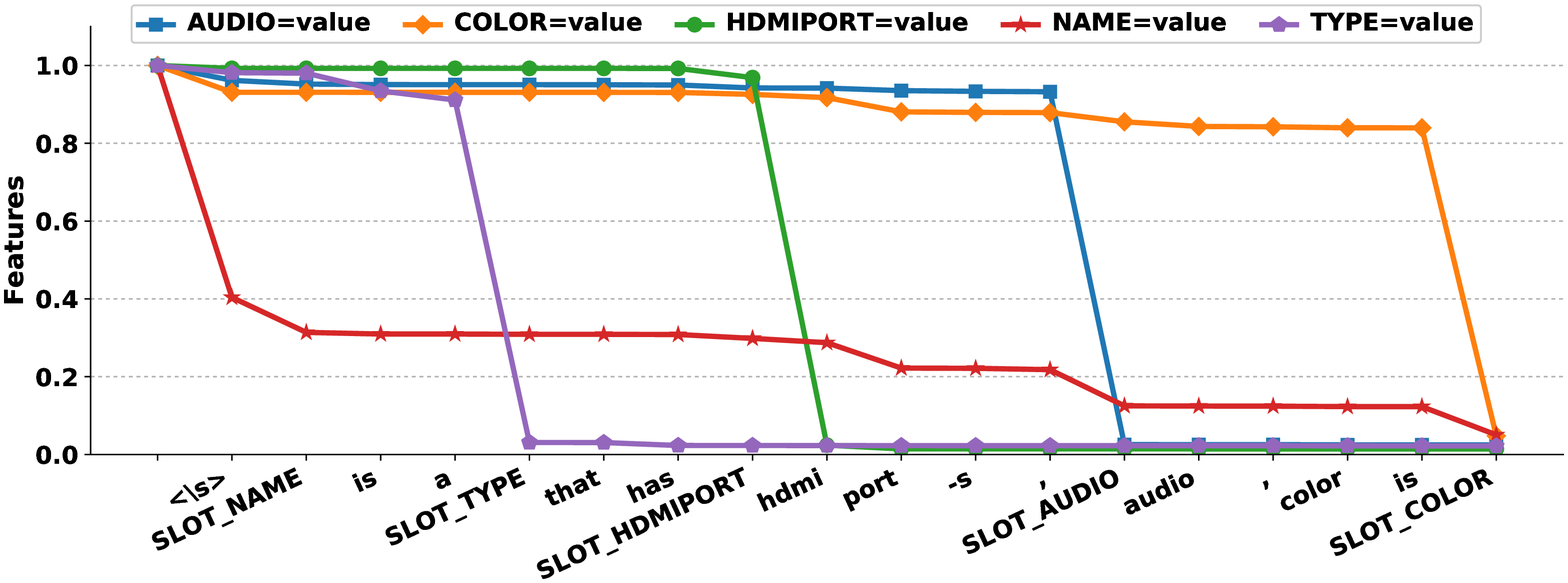}
      \subcaption{An example from the TV domain.}
      \label{fig:Sample} 
    \end{subfigure}
\caption{Example showing how RALSTM drives down the DA feature value vector $\textbf{s}$ step-by-step, in which the model generally shows its ability to detect words and phases describing a corresponding slot-value pair.}   
\label{fig:DAVectoControlling}
\end{figure}

All these prove the importance of the proposed components: the Refinement cell in aggregating and selecting the attentive information, and the Adjustment cell in controlling the feature vector (see Examples in Figure \ref{fig:DAVectoControlling}).
\subsubsection*{General Models}
Figure \ref{fig:general-models} shows a comparison performance of general models as described in Section \ref{subsec:experimental-setups}. The results are consistent with the Figure \ref{fig:laptop-performances}, in which the \textit{RALSTM} has better performance than the \textit{Enc-Dec} and \textit{SCLSTM} on all domains in terms of the BLEU and the ERR scores, while the \textit{Enc-Dec} has difficulties in reducing the ERR score. This indicates the relevant contribution of the proposed component Refinement and Adjustment cells to the original ARED architecture, in which the Refinement with attentional gating can effectively select and aggregate the information before putting them into the traditional LSTM cell, while the Adjustment with gating DA vector can effectively control the information flow during generation.
\subsubsection*{Adaptation Models}
Figure \ref{fig:adaptation-models} shows domain scalability of the three models in which the models were first trained on the merging out-of-domain Restaurant and Hotel datasets, then fine tuned the parameters with varied amount of in-domain training data (laptop domain). The \textit{RALSTM} model outperforms the previous model in both cases where the sufficient in-domain data is used (as in Figure \ref{fig:adaptation-models}-\textit{left}) and the limited in-domain data is used (Figure \ref{fig:adaptation-models}-\textit{right}). The Figure \ref{fig:adaptation-models}-\textit{right} also indicates that the \textit{RALSTM} model can adapt to a new, unseen domain faster than the previous models.

\section{Conclusion and Future Work}\label{sec:conclusion}
We present an extension of ARED model, in which an RALSTM component is introduced to select and aggregate semantic elements produced by the Encoder, and to generate the required sentence. We assessed the proposed models on four NLG domains and compared to the state-of-the-art generators. The proposed models empirically show consistent improvement over the previous methods in both the BLEU and ERR evaluation metrics. The proposed models also show an ability to extend to a new, unseen domain no matter how much the in-domain training data was fed. In the future, it would be interesting to apply the proposed model to other tasks that can be modeled based on the encoder-decoder architecture, i.e., image captioning, reading comprehension, and machine translation.
\bibliography{acl2017}
\bibliographystyle{acl_natbib}

\end{document}